\documentclass[letterpaper,twocolumn,fleqn]{article} 

\usepackage{ist}
\usepackage{hyperref}
\usepackage{algorithm2e}
\usepackage{caption}
\usepackage{subcaption}
\usepackage{float}
\usepackage{etoolbox}
\usepackage{amssymb}
\usepackage{bbold}
\usepackage{soul}

\title{Boosting Image Forgery Detection using Resampling Features and Copy-move Analysis}

\author{Tajuddin Manhar Mohammed; Mayachitra Inc., Santa Barbara, California, USA\\
Jason Bunk; Mayachitra Inc., Santa Barbara, California, USA\\
Lakshmanan Nataraj; Mayachitra Inc., Santa Barbara, California, USA\\
Jawadul H. Bappy; University of California, Riverside, California, USA\\
Arjuna Flenner; Naval Air Warfare Center Weapons Division, China Lake, California, USA\\
B. S. Manjunath; Mayachitra Inc., Santa Barbara, California, USA\\ Shivkumar Chandrasekaran;  Mayachitra Inc., Santa Barbara, California, USA\\
Amit K. Roy-Chowdhury; University of California, Riverside, California, USA\\
Lawrence Peterson; Naval Air Warfare Center Weapons Division, China Lake, California, USA}

\date{} 

\begin{document} 



\maketitle 

\thispagestyle{empty} 


\begin{abstract}

Realistic image forgeries involve a combination of splicing, resampling, cloning, region removal and other methods. While resampling detection algorithms are effective in detecting splicing and resampling, copy-move detection algorithms excel in detecting cloning and region removal. In this paper, we combine these complementary approaches in a way that boosts the overall accuracy of image manipulation detection. We use the copy-move detection method as a pre-filtering step and pass those images that are classified as untampered to a deep learning based resampling detection framework. Experimental results on various datasets including the 2017 NIST Nimble Challenge Evaluation dataset comprising nearly 10,000 pristine and tampered images shows that there is a consistent increase of 8\%-10\% in detection rates
, when copy-move algorithm is combined with different resampling detection algorithms.

\end{abstract}

\section{Introduction}
\label{sec:intro}

\begin{figure*}[t]
\centering
\captionsetup{justification=centering}
\includegraphics[scale=0.68]{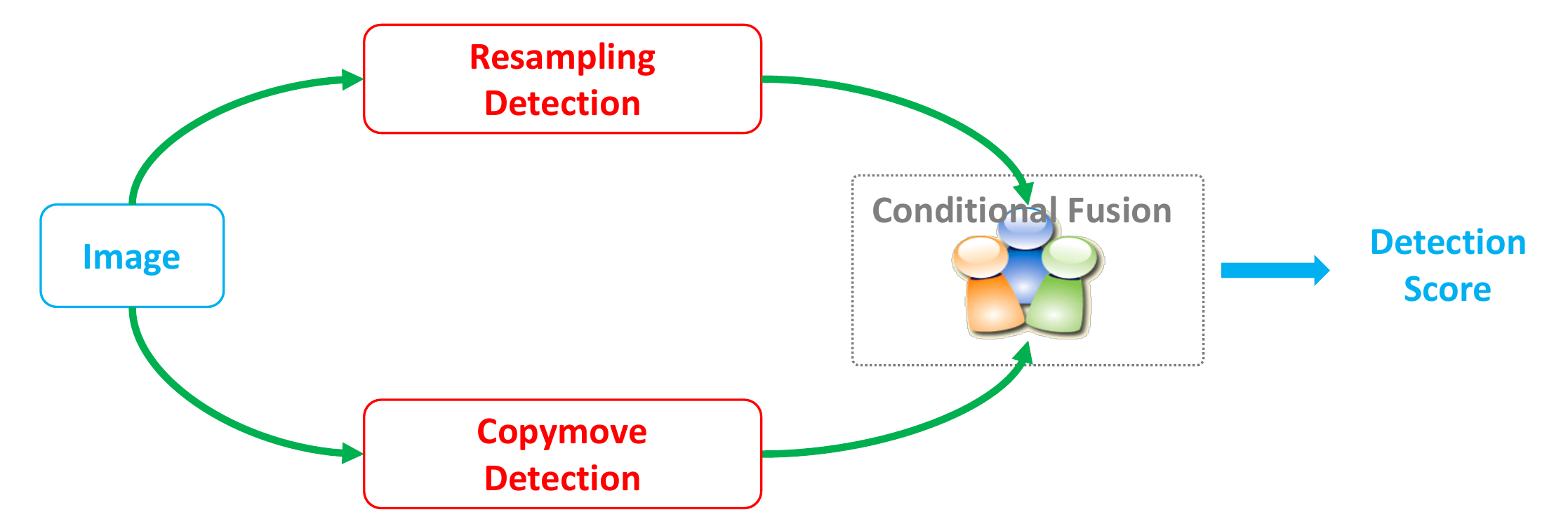}
\caption{Block Schematic of our proposed approach } 
\label{fig:over}
\vspace{-5pt}
\end{figure*}

Fake images are becoming a growing threat to information reliability. 
With the ubiquitous availability of various powerful image editing software tools and smartphone apps such as Photoshop, GIMP, Snapseed and Pixlr, it has become very trivial to manipulate digital images.
The field of Digital Image Forensics aims to develop tools that can identify the authenticity of digital images and localize regions in an image which have been tampered with. 

There are many types of image forgeries such as splicing objects from one image to another, removing objects or regions from images, creating copies of objects in the same image, and more. 
To detect these forgeries, researchers have proposed methods based on several techniques such as JPEG compression artifacts, resampling detection, lighting artifacts, noise inconsistencies, camera sensor noise, and many more. 
However, most techniques in literature focus on a specific type of manipulation or a groups of similar tamper operations. 
In realistic scenarios, a host of operations are applied when creating tampered images.
For example, when an object is spliced onto an image, it is often accompanied by other operations such as scaling, rotation, smoothing, contrast enhancement, and more.
Very few studies address these challenging scenarios with the aid of Image Forensics challenges and competitions such as IEEE Image Forensics challenge~\cite{2013-ifc-challenge} and the recent NIST Nimble Media Forensics challenge~\cite{nist2017}.
These competitions try to mimic a realistic scenario and contain a large number of doctored images which involves several types of image manipulations. 
In order to detect the tampered images, a single detection method will not be sufficient to identify the different types of manipulations.
In this paper, we demonstrate the importance of combining forgery detection algorithms, especially when the features are complementary, to boost the image manipulation detection rates.
We propose a simple method to identify realistic forgeries by fusing two complementary approaches: resampling detection and copy-move detection.
Our experimental results show the approach is promising and achieves an increase in detection rates.

Image forgeries are usually created by splicing a portion of an image onto some other image. 
In the case of splicing or object removal, the tampered region is often scaled or rotated to make it proportional to the neighboring untampered area. 
This creates resampling of the image grid and detection of resampling indicates evidence of image manipulation. 
Several techniques have been proposed to detect resampling in digital images~\cite{popescu-farid-resampling, prasad2006resampling, second-diff-method, babak-radon, kirchner-local, feng2012normalized, ryu2014estimation}. 
Similarly, copy-move forgeries are common, where a part of the image is copied and pasted on another part generally to conceal unwanted portions of the image.  
Detection of these copied parts indicates evidence of tampering~\cite{fridrich2003detection,popescu-farid-tr-04,mahdian2007detection,bayram2009efficient,christlein2012evaluation,cao2012robust,cozzolino2015efficient,rao2016deep}.

In this paper, we combine our previous work on resampling forgery detection~\cite{bunk2017detection} with a dense-field based copy-move forgery detection method developed by Cozzolino et al.~\cite{cozzolino2015efficient} to assign a manipulation confidence score. 
We demonstrate that our algorithm is effective at detecting many different types of image tampering that can be used to verify the authenticity of digital images. 
In~\cite{bunk2017detection}, we designed a detector based on Radon transform and deep learning. 
The detector found image artifacts imposed by classic upsampling, downsampling, clockwise and counter clockwise rotations, and shearing methods.
We combined these five different resampling detectors with a JPEG compression detector and for each of the six detectors we output a heatmap which indicates the regions of resampling anomalies.
The generated heatmaps were smoothed to localize the detection and determine the detection score. 
In this work, we combine the above approach with a copy-move forgery detector~\cite{cozzolino2015efficient}.
Our experiments demonstrate that the resampling features are complementary to the copy-move forgery detection features and combining them indeed increases the image manipulation detection rates. 

The rest of the paper is organized as follows.
In the next section, we describe the related work in image forgery detection. 
After that, we describe the methodology, explain the results, and provide a discussion for combining such complementary methods. 
Finally, we conclude the paper and outline our future work.

\section{Related Work}
The image forensics field recognizes many different categories of image forgeries with copy-move, splicing, and  object removal as some of the most popular.  In many of these forgeries, resampling is a necessary work-flow element.  Due to the diversity of image tampering methods, there are numerous techniques to identify if an image has been manipulated and we briefly review some of the techniques below with a special emphasis on techniques that overlap with our method.

In computer vision, deep learning showed outstanding performance in different visual recognition tasks such as image classification~\cite{zhou2014learning}, and semantic segmentation~\cite{long2015fully}.  For this reason, there has been a growing interest to detect image manipulation by applying different computer vision and deep-learning algorithms~\cite{bayar2016deep, bayar2017design,rao2016deep,bappy2017exploiting}.  In~\cite{long2015fully}, two fully convolution layers have been exploited to segment different objects in an image. The segmentation task has been further improved in \cite{zheng2015conditional, badrinarayanan2017segnet}. These models extract hierarchical features to represent the visual concept, which is useful in object segmentation. Since, the manipulation does not exhibit any visual change with respect to genuine images, these models often do not perform well in segmenting manipulated regions. 

Other deep learning methods include detection of generic manipulations~\cite{bayar2016deep, bayar2017design}, resampling~\cite{bayar2017resampling}, splicing~\cite{rao2016deep} and bootleg~\cite{buccoli2014unsupervised}.   In~\cite{qian2015deep}, the authors propose Gaussian-Neuron CNN (GNCNN) for steganalysis detection.
A deep learning approach to identify facial retouching was proposed in~\cite{bharati2016detecting}.
In~\cite{zhang2016image}, image region forgery detection has been performed using a stacked auto-encoder model. 
In~\cite{bayar2016deep},  a new constrained convolutional layer is proposed to learn the manipulated features  from an image. In our previous work~\cite{bunk2017detection}, a unique network exploiting convolution layers along with LSTM network was presented.

\subsection{Resampling Forgery Detection}
Resampling an image requires an interpolation method and linear or cubic interpolations are very popular and this fact was exploited by the authors of~\cite{popescu-farid-resampling}.  They implemented an Expectation-Maximization (EM) algorithm to detect  periodic correlations introduced by interpolation.   However, the periodic JPEG blocking artifacts also introduce periodic patterns that confuses their resampling detector.
The variance of the second difference operator was used by ~\cite{second-diff-method} to detect resampling on images that are scaled using linear or cubic interpolations. Their method is most efficient at detecting up-scaling and it is very robust to JPEG compression with detection possible even at very low quality factors (QF). Downscaled images can be detected but not as robustly as upscaled images.  In ~\cite{babak-radon}, the Radon transform and a derivative filter was exploited to improve the quality of the results and to address other forms of resampling.  In~\cite{ kirchner-local}, a simpler method than~\cite{popescu-farid-resampling} was derived by using a linear predictor residue instead of the computationally expensive EM algorithm.  This resampling body of work motivated us in~\cite{bunk2017detection}, where we combined the linear predictor strategy with deep learning based models in order to detect tampered image patches.

Other resampling detectors include~\cite{ryu2014estimation, Nataraj10-345,Nataraj09-331, feng2011energy,feng2012normalized, golestaneh2014algorithm, kwon2015efficient}.  In~\cite{ryu2014estimation}, periodic properties of interpolation were found using the second-derivative of the image and these properties were used for detecting image manipulation.  Resampling on JPEG compressed images was detected in~\cite{Nataraj10-345,Nataraj09-331} by adding noise before passing the image through the resampling detector and they showed that noise addition improved resampling detection. 
A normalized energy feature was implemented in~\cite{feng2011energy,feng2012normalized} and a support vector machine (SVM) was subsequently used to classify resampled images. 
Furthermore, recent approaches to reduce the effects of JPEG artifacts were developed in~\cite{golestaneh2014algorithm, kwon2015efficient}.

\subsection{Copy-Move Forgery Detection}
Copy-move forgery is a specific type of image tampering, where a part of the image is copied and pasted on to another part of the same image.
For copy-move forgeries, a common approach is to match image features within the image.  In order to detect copy-move forgeries, an image is first divided into overlapping blocks and some sort of distance measure or correlation is used to determine blocks that have been cloned. For example, in~\cite{amerini2011sift}, copy-move forgeries were detected using SIFT features.  Many similar methods to detect copy-move have been proposed~\cite{li2015segmentation,kakar2012exposing, jaberi2014accurate,al2013passive}.

Another strategy to detect copy move forgeries is to match a transformation of image regions rather than image regions themselves.  In \cite{fridrich2003detection}, Fridrich et al. used DCT coefficients of image regions to find duplicate DCT blocks while Popsecu and Farid used PCA~\cite{popescu-farid-tr-04} to detect duplicated regions and Mahdian and Saic use a combination of blur invariant moments and PCA~\cite{mahdian2007detection}.  A matching image regions to detect copy-move forgeries becomes more difficult if the moved region undergoes some transformation such as scaling that makes region matching difficult. Bayram et al.~\cite{bayram2009efficient} addresses this issue by using a combination of Fourier Mellin transforms, which are invariant to rotation, scale and translation, and Bloom filters. 
Another issue in locating copy-move forgeries is the computational time to find matching patches.

In this paper, we use the work of Cozzolino et al.~\cite{cozzolino2015efficient},  where a patch-match algorithm is used to efficiently compute an approximate nearest neighbor field over an image.
They added robustness to their algorithm by using invariant features such as Circular Harmonic transforms and show that they can detect duplicated blocks that have undergone geometrical transformations and then perform key-point matching.


\section{Methodology}

\begin{figure}[!htbp]
\centering
\captionsetup{justification=centering}
\includegraphics[scale=0.38]{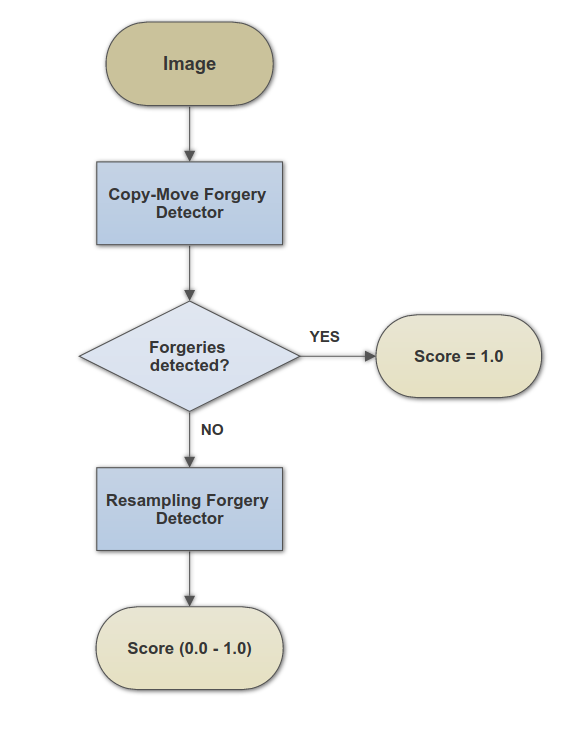}
\caption{Flowchart for our proposed algorithm} 
\label{fig:ovw}
\end{figure}

\begin{figure*}[!htbp]
\centering
\captionsetup{justification=centering}
\includegraphics[scale=0.3]{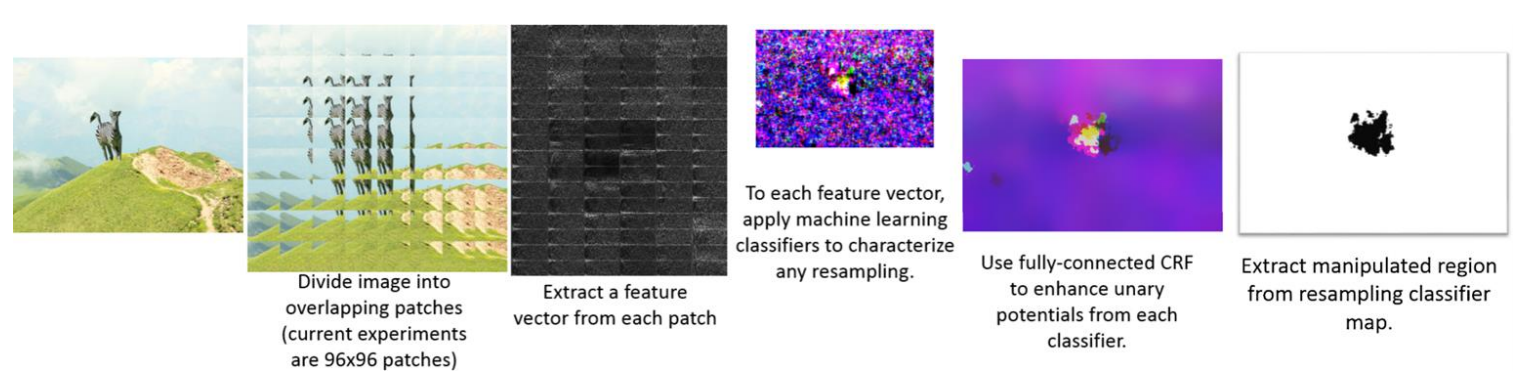}
\caption{An end-to-end framework to detect and localize digital manipulations based on resampling anomalies} 
\label{fig:resamp}
\end{figure*}

Our method to get a confidence score to determine whether a digital image is manipulated involves two primary steps summarized in Fig.~\ref{fig:ovw}. First, we pass the digital image through a copy-move forgery detector to detect clone forgeries. Second, we use the resampling features to detect manipulations if the former detection algorithm does not detect a forgery. In other words, the copy-move forgery detection will be used as a ``pre-filtering'' step to improve the precision of the resampling forgery detection, and indeed boost the image manipulation detection rates. The outline for both the copy-move and resampling forgery detection algorithms will be presented here.

\subsection{Dense-field based Copy-Move Forgery Detection}

\begin{figure}[H]
\centering
\captionsetup{justification=centering}
\includegraphics[scale=0.68]{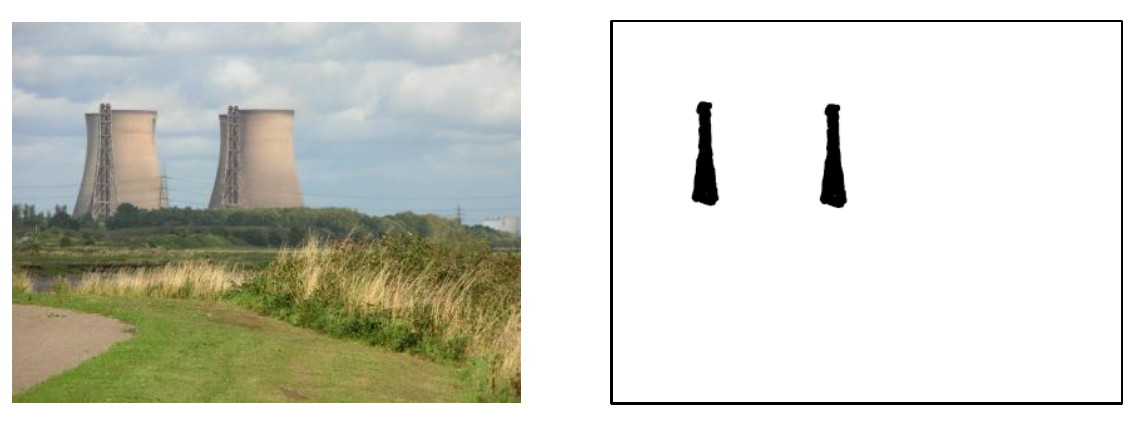}
\caption{Example of a copy-move forged image and the dense-field based copy-move forgery detection based on~\cite{cozzolino2015efficient} } 
\label{fig:cpmv}
\end{figure}

Cozzolino et al.~\cite{cozzolino2015efficient} proposed a fast and accurate CMF detection algorithm based on a modified Patch-Match algorithm~\cite{barnes2009patchmatch}, for rotation-invariant and scale-invariant forgery detection. The Patch-Match algorithm was used to compute efficiently a high-quality approximate nearest neighbor field for the whole image. They replaced the use of RGB pixel values in the Patch-Match algorithm to scale and rotation features. These features include Zernike Moments (ZM)~\cite{teague1980image}, Polar Cosine Transform (PCT)~\cite{yap2010two} and Fourier-Mellin Transform (FMT)~\cite{sheng1986experiments}. Then they applied a post-processing procedure to remove instances of false matching. We leave the details to their work~\cite{cozzolino2015efficient}. Although, the efficiency of their algorithm is reduced for blurred images, their method is shown to be robust to translation, rotation, moderate scaling and post-processing methods like noise addition and JPEG compression.

\subsection{Deep Learning based Resampling Forgery Detection}

\begin{figure}[H]
\centering
\captionsetup{justification=centering}
\includegraphics[scale=0.68]{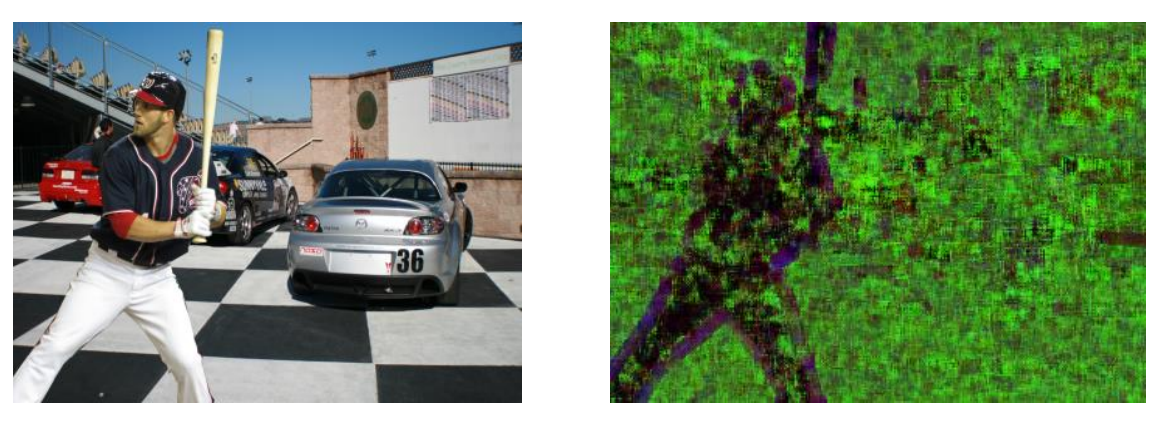}
\caption{Example of a spliced image and the Radon transform based resampling detection heatmap based on~\cite{bunk2017detection} } 
\label{fig:res}
\end{figure}

In our previous work~\cite{bunk2017detection}, we proposed an end-to-end Deep Learning framework (Fig.~\ref{fig:resamp}) to detect and localize manipulations based on resampling anomalies in a digital image. We first extract image patches, apply a 3x3 Laplacian filter to these patches and then compute the linear predictive error. To look for periodic correlations, we apply Radon transform to accumulate errors along various angles of projection, and then compute the Fast Fourier Transform (FFT) to find the periodicity. This Radon Transform based technique was inspired from~\cite{mahdian2008blind}. Thus, we get our desired resampling features. The second step is to characterize any resampling detected in the patch. We train a set of six binary classifiers that check for different types of resampling: JPEG quality with a threshold above or below 85, upsampling, downsampling, rotation clockwise, rotation counterclockwise, and shearing. To train a binary classification model for each task, we build a dataset of about 100,000 patches extracted from about 8,000 images from two publicly available datasets of uncompressed images, UCID~\cite{schaefer2003ucid} and RAISE~\cite{dang2015raise} datasets. Some of the patches are transformed with a set of randomly generated parameters, such as multiple JPEG compressions and affine transformations, but one-half of the dataset included the transformation specified, and the other half did not. The classifiers are not mutually exclusive and are trained individually. The best performing binary classifier we found for this task was an artificial neural network or a multi-layer perceptron with two hidden layers. The filtering in the third step uses bilateral filtering, which is commonly used in semantic segmentation when fusing adjacent noisy local patch-based classifiers. The final step is to obtain an image-level detection score using a median metric and a mask that shows manipulated regions from resampling heatmaps using median filtering, Otsu thresholding and Random Walker segmentation. The entire algorithm is summarized in Fig.~\ref{fig:resamp}.


\section{Results and Discussion}

In this section, we describe the dataset and the metric used to evaluate the performance of our proposed model. After this, we present and explain our results and provide a discussion for combining such complementary methods.

\subsection{2017 Nimble Evaluation Dataset}

The 2017 NIST Nimble Challenge evaluation dataset~\cite{nist2017} comprises around 10,000 images with numerous type of local/global manipulations including the ones where anti-forensic algorithms were used to hide the trivial manipulations. As part of the challenge, the participating teams were also provided different development datasets with ground-truth and other relevant metadata information that could be used to train and test the models. 

\subsection{Area Under the ROC Curve (AUC)}

The receiver operating characteristic (ROC) curve is a graphical plot that demonstrates the ability of a binary classification system as its discrimination threshold is varied. Macmillan and Creelman~\cite{macmillan2004detection} provide detailed information about ROC curves for detection system evaluation. AUC quantifies the overall ability of the system to discriminate between two classes. A system no better at identifying true positives than random guessing has an AUC of 0.5. A perfect system (no false positives or false negatives) has an AUC of 1.0. The AUC value of a system output has a value between 0 and 1.0.

\subsection{Results on the dataset}
\begin{figure*}[!htbp]-    \centering
    \begin{subfigure}[t]{0.475\textwidth}
        \centering
        \includegraphics[width=1\columnwidth]{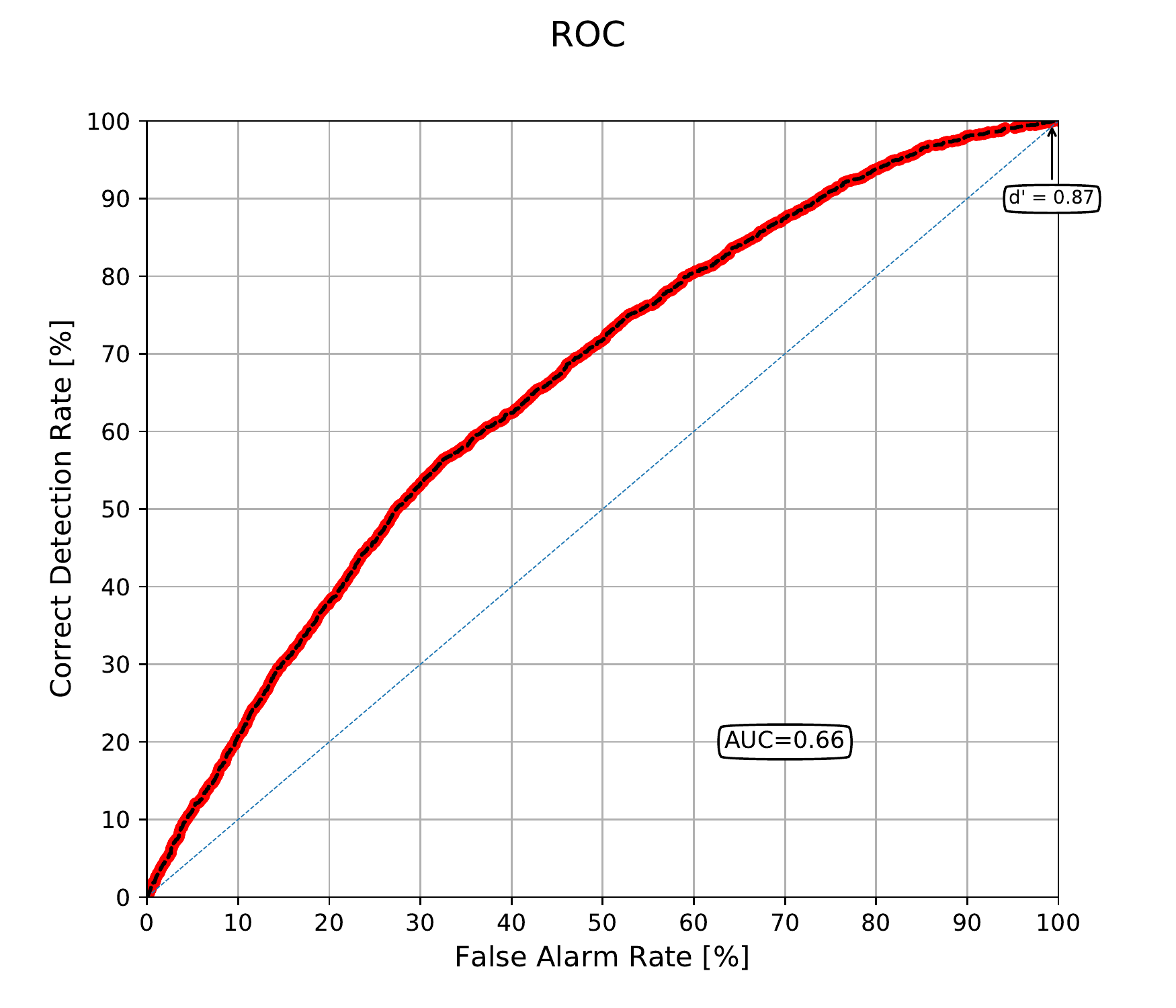}
        \caption{}
    \end{subfigure}%
    ~
    \begin{subfigure}[t]{0.475\textwidth}
        \centering
        \includegraphics[width=1\columnwidth]{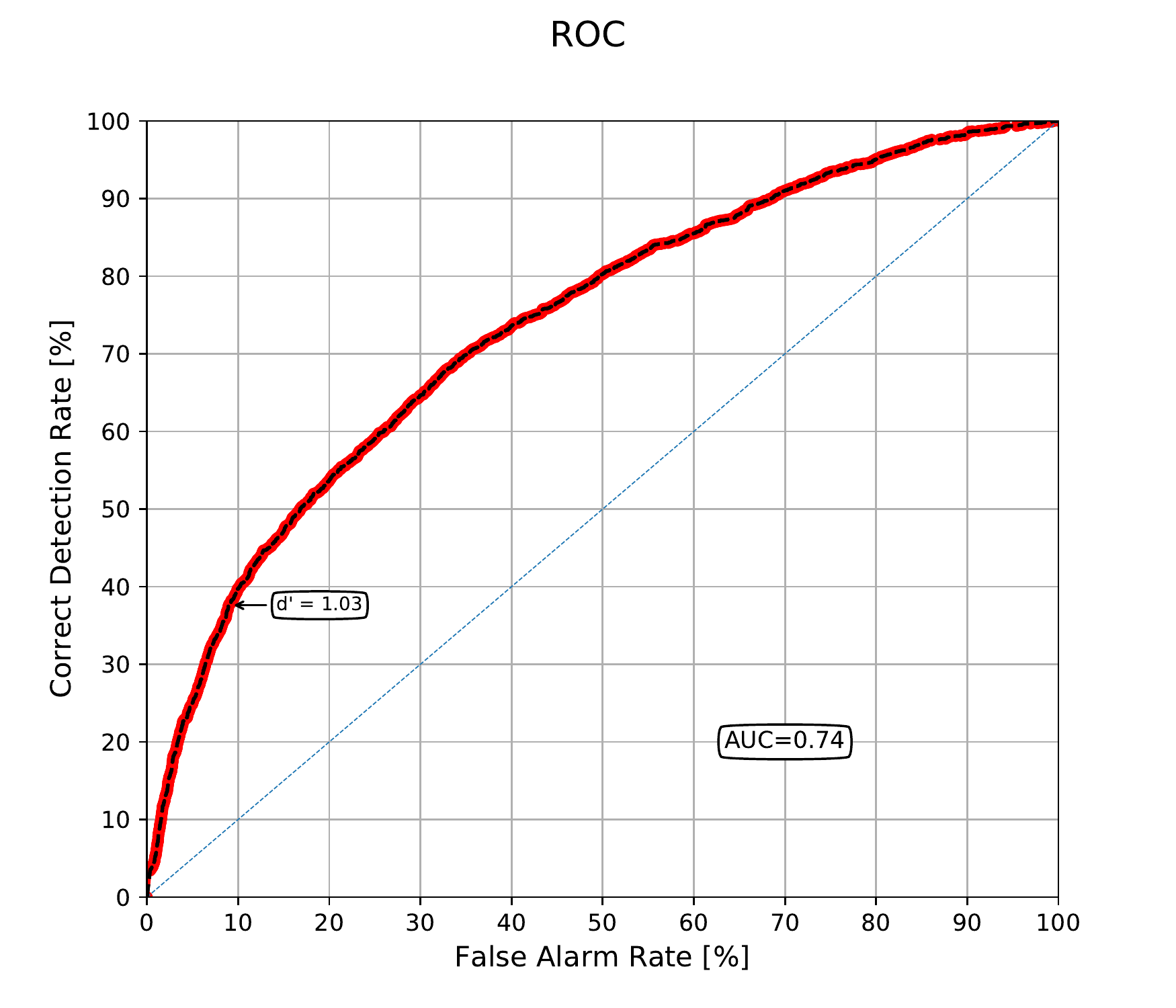}
        \caption{}
    \end{subfigure}
    \vspace{-2pt}
    \caption{ROC curves and corresponding AUC scores of our (a) Resampling forgery detector~\cite{bunk2017detection} and (b) Proposed method on the full 2017 Nimble evaluation dataset}
    \label{fig:fullset}
    \vspace{-5pt}
\end{figure*}

Our proposed solution that combined both copy-move forgery detection and resampling forgery detection models obtained an overall AUC score of 0.74 on the 2017 Nimble Evaluation Dataset. 
As shown in Fig.~\ref{fig:fullset}, there is a 8\% boost in AUC scores (0.66 to 0.74) when we compare our proposed model to that of the resampling forgery detector~\cite{bunk2017detection}, when used independently.
Individually, the copy-move method was able to get an AUC of 0.64. 

\subsection{Error Analysis}

\begin{figure*}[!htbp]
    \centering
    \begin{subfigure}[t]{0.475\textwidth}
        \centering
        \includegraphics[width=1\columnwidth]{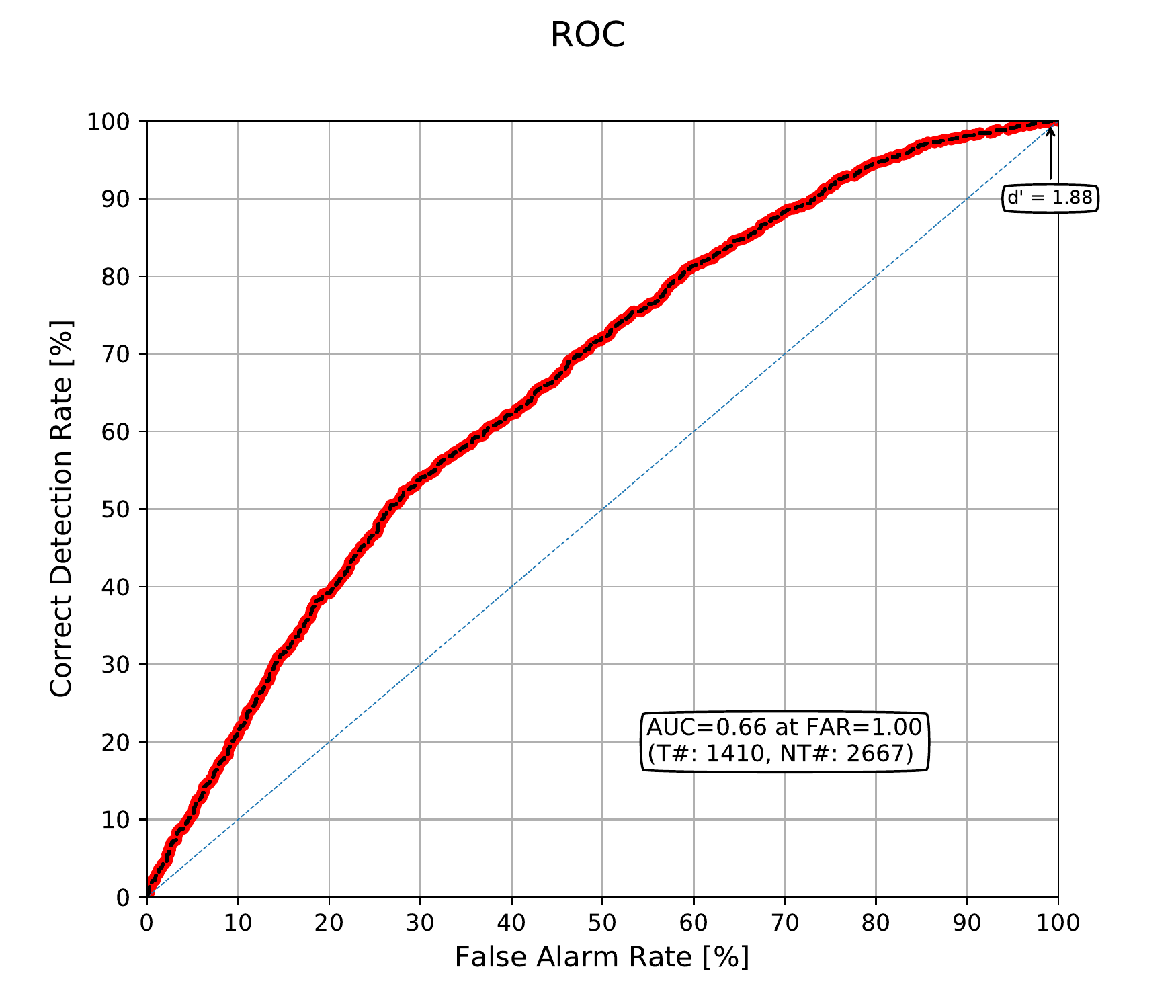}
        \caption{}
    \end{subfigure}%
    ~
    \begin{subfigure}[t]{0.475\textwidth}
        \centering
        \includegraphics[width=1\columnwidth]{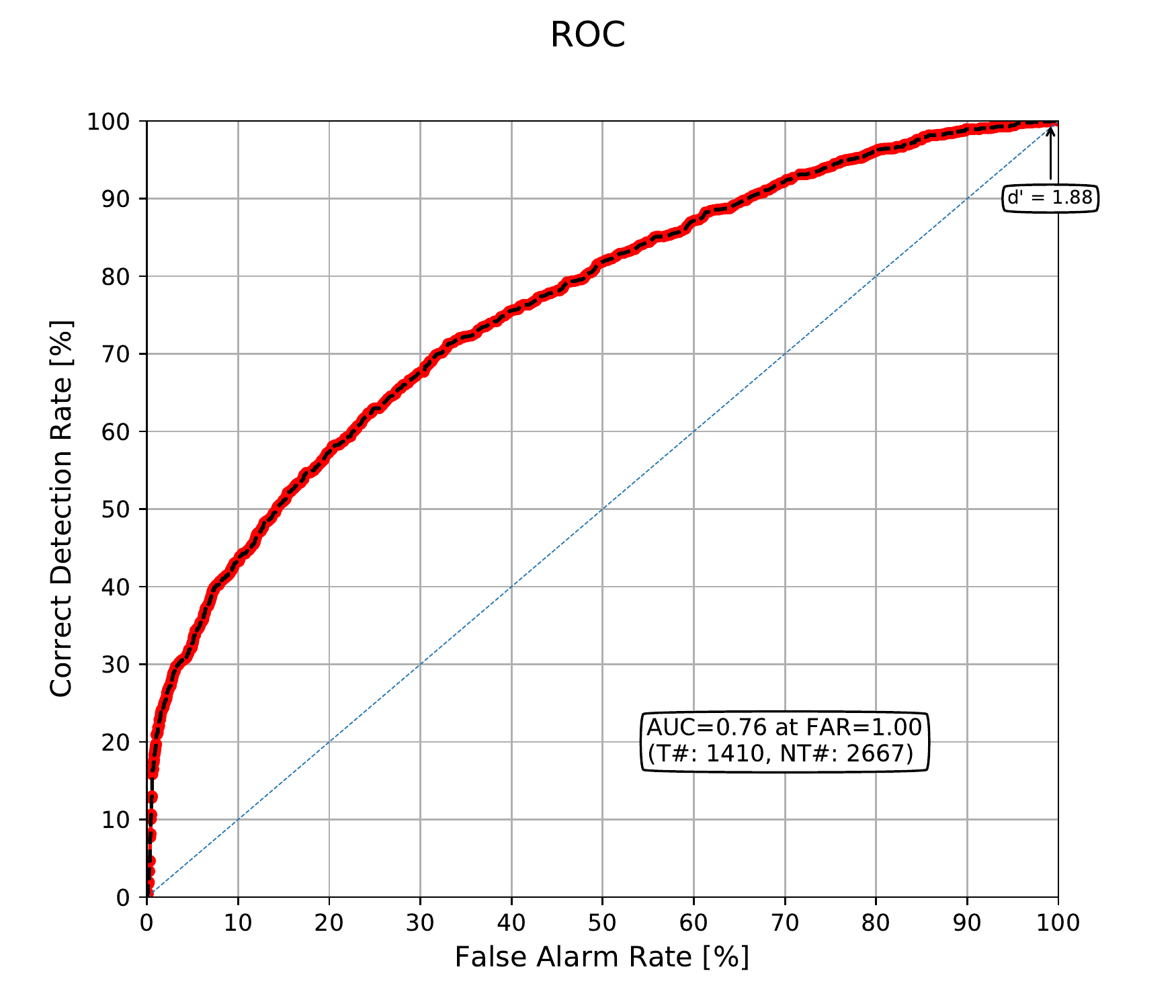}
        \caption{}
    \end{subfigure}
    \vspace{-2pt}
    \caption{ROC curves and corresponding AUC scores of our (a) Resampling forgery detector~\cite{bunk2017detection} and (b) Proposed method on the 1/3rd subset of the 2017 Nimble evaluation dataset}
    \label{fig:thirds}
    \vspace{-5pt}
\end{figure*}

We also performed an error analysis on the 1/3rd of the evaluation dataset for which the ground-truth information was provided by NIST, after the challenge. There are 4,077 images in this subset of the data, out of which, 1,410 are manipulated and the rest 2,667 are un-manipulated. As shown in Fig.~\ref{fig:thirds}, a boost of 10\% in the AUC scores (0.66 to 0.76) is observed, as expected. The pre-filtering step which is the copy-move forgery detection correctly identified 549 out of 1,410 manipulated images and missed the remaining 861 images.  But, 532 out of these 861 images (62\%) were flagged manipulated by our resampling forgery detector. This is reflected in the 10\% boost in AUC scores. For this analysis, the optimal threshold for the resampling detector was selected such that the \textit{true positive rate} from the ROC curve was high and the \textit{false positive rate} was low. 

Similarly, there was a consistent 8\%-10\% improvement in AUC scores for other development datasets provided by NIST. These results demonstrate the importance of combining complementary methods of forgery detection (in this case, copy-move forgery detection and resampling forgery detection) to improve the manipulation detection rates.


\section{Conclusion and Future Work}


In this paper, we described a technique to determine a manipulation score for a digital image based on the resampling features~\cite{bunk2017detection} and copy-move features~\cite{cozzolino2015efficient}. We demonstrated the complementary nature of these features and the importance of using copy-move detection algorithm as a pre-filtering step to resampling forgery detection in order to boost the image manipulation detection rates. Experimental results show that our proposed approach increases the AUC scores consistently by 8\%-10\% for various datasets. 

As we used the work of Cozzolino et al.~\cite{cozzolino2015efficient} for the copy-move pre-filtering step, there is a room for improvement as the copy-move detection algorithm, even-though efficient generates moderate number of false-positives. 
And also, a better way to combine these works, rather than using copy-move forgery detection as a pre-filtering step can be exploited. 

\section{Acknowledgments} 

This research was developed with funding from the Defense Advanced Research Projects Agency (DARPA).
The views, opinions and/or findings expressed are those of the author and should not be interpreted as representing the official views or policies of the Department of Defense or the U.S. Government. 
The paper is approved for public release, distribution unlimited.


{\small
	\bibliographystyle{spiejour}
	\bibliography{resamp-boost.bbl}
}


\begin{biography}

\textbf{Tajuddin Manhar Mohammed} received his B.Tech (Hons.) degree in Electrical Engineering from Indian Institute of Technology (IIT), Hyderabad, India in 2015 and his M.S. degree in Electrical and Computer Engineering from University of California Santa Barbara (UCSB), Santa Barbara, CA in 2016. After obtaining his Masters degree, he obtained a job as a Research Staff Member for Mayachitra Inc., Santa Barbara, CA. His recent research efforts include developing deep learning and computer vision techniques for image forensics and cyber security.

\textbf{Jason Bunk} received his B.S. degree Computational Physics from the University of California, San Diego in 2015, and his M.S. degree in Electrical and Computer Engineering from the University of California, Santa Barbara in 2016. 
He is currently a Research Staff Member at Mayachitra Inc., Santa Barbara, CA. 
His recent research efforts include applying deep learning techniques to media forensics, and active learning with neural networks for video activity recognition.

\textbf{Lakshmanan Nataraj} received his B.E degree in Electronics and Communications Engineering from Anna University in 2007, and his Ph.D. degree in the Electrical and Computer Engineering from the University of California, Santa Barbara in 2015. 
He is currently a Research Staff Member at Mayachitra Inc., Santa Barbara, CA. 
His research interests include malware analysis and image forensics. 

\textbf{Jawadul H. Bappy} received his B.S. degree in Electrical and Electronic Engineering from the Bangladesh University of Engineering and Technology, Dhaka in 2012. 
He is currently pursuing his Ph.D. degree in Electrical and Computer Engineering at University of California, Riverside.
His main research interests include wide area scene analysis, scene understanding, object recognition and machine learning.

\textbf{Arjuna Flenner} received his Ph.D. in Physics at the University of Missouri-Columbia located in Columbia MO in the year 2004. His major emphasis was mathematical Physics. After obtaining his Ph.D., Arjuna Flenner obtained a job as a research physicist for NAVAIR at China Lake CA. He won the 2013 Dr. Delores M. Etter Navy Scientist and Engineer award for his work on Machine Learning. 

\textbf{B. S. Manjunath} received his Ph.D. degree
in Electrical Engineering from the University of
Southern California, Los Angeles, CA, USA,
in 1991. He is currently a Professor in Electrical and
Computer Engineering department at University of
California at Santa Barbara, Santa Barbara, CA,
USA, where he directs the Center for Multi-Modal
Big Data Science and Healthcare. He has authored or
co-authored about 300 peer-reviewed articles. 

\textbf{Shivkumar Chandrasekaran} received his M.Sc. (Hons.)
degree in physics from the Birla Institute of Technology and Science (BITS), Pilani, India, in 1987,
and his Ph.D. degree in Computer Science from Yale
University, New Haven, CT, in 1994.
He was a Visiting Instructor at North Carolina
State University, Raleigh, in the Mathematics Department,
before joining the Electrical and Computer
Engineering Department, University of California,
Santa Barbara, where he is currently a Professor. His
research interests are in Computational Mathematics.

\textbf{Lawrence A. Peterson} has extensive research and development experience as a Civilian Engineer / Scientist with the U. S. Navy for over 37 years. He currently heads the  Image and Signal Processing Branch in the Physics and Computational Sciences Division,  Research Directorate at the Naval Air Warfare Center, Weapons Division (NAWCWD), China Lake, CA.  


\textbf{Amit K. Roy-Chowdhury} received his Ph.D. degree in
Electrical Engineering from University of
Maryland, College Park.
He is a Professor of Electrical Engineering
and a Cooperating Faculty in the Department
of Computer Science, University of California,
Riverside. 
His broad research interests include the areas of
image processing, computer vision, and video communications and statistical methods for signal analysis. 
He has been on the organizing and program committees
of multiple computer vision and image processing conferences and is serving on the editorial boards of multiple journals.

\end{biography}

\end{document}